\pdfoutput=1

\documentclass[11pt]{article}

 \usepackage[]{emnlp2021}

\usepackage{times}
\usepackage{latexsym}

\usepackage[T1]{fontenc}

\usepackage[utf8]{inputenc}

 \usepackage{amsmath}
\usepackage{microtype}
\usepackage{graphicx}
\usepackage{xcolor}
\usepackage{multirow}
\usepackage{booktabs}
\usepackage{setspace}

\title{Back to the Basics: A Quantitative Analysis of Statistical and Graph-Based Term Weighting Schemes for Keyword Extraction}

\author{Asahi Ushio \and Federico Liberatore \and Jose Camacho-Collados\\
  School of Computer Science and Informatics \\ Cardiff University, United Kingdom\\
  \texttt{\{UshioA,LiberatoreF,CamachoColladosJ\}@cardiff.ac.uk}
  \\}

\begin{document}
\maketitle


\begin{abstract}
 Term weighting schemes are widely used in Natural Language Processing and Information Retrieval. In particular, term weighting is the basis for keyword extraction. However, there are relatively few evaluation studies that shed light about the strengths and shortcomings of each weighting scheme. In fact, in most cases researchers and practitioners resort to the well-known tf-idf as default, despite the existence of other suitable alternatives, including graph-based models. In this paper, we perform an exhaustive and large-scale empirical comparison of both statistical and graph-based term weighting methods in the context of keyword extraction. Our analysis reveals some interesting findings such as the advantages of the less-known lexical specificity with respect to tf-idf, or the qualitative differences between statistical and graph-based methods. Finally, based on our findings we discuss and devise some suggestions for practitioners.\footnote{Source code to reproduce our experimental results, including a keyword extraction library, are available in the following repository:  \url{https://github.com/asahi417/kex}}

\end{abstract}

\section{Introduction}
\label{sec:introduction}


Keyword extraction has been an essential task in many scientific fields as a first step to extract relevant terms from text corpora. Despite the simplicity of the task, it still poses practical problems, and often researchers resort to simple but reliable techniques such as tf-idf \cite{jones1972statistical}. In turn, term weighting schemes such as tf-idf paved the way for developing large-scale Information Retrieval (IR) systems \cite{ramos2003using,wu2008interpreting}. Its simple formulation is still widely used nowadays, not only for keyword extraction but also as an important component in IR \cite{jabri2018ranking,marcos2020information} and Natural Language Processing (NLP) tasks \cite{riedel2017simple,arroyo2019unsupervised}. 

While there exist supervised and neural techniques \cite{lahiri2017keyword,xiong-etal-2019-open,sun2020joint}, as well as ensembles of unsupervised methods \cite{campos2020yake,tang2020improved} that can provide competitive performance, in this paper we go back to the basics and analyze in detail the single components of unsupervised methods for keyword extraction. In fact, it is still common to rely on unsupervised methods for keyword extraction given their versatility and the lack of training sets in specialized domains.

In order to fill this gap, in this paper we perform an extensive analysis of single unsupervised keyword extraction techniques in a wide range of settings and datasets. 
To the best of our knowledge, this is the first large-scale empirical evaluation performed across base statistical and graphical keyword extraction methods. Our analysis sheds light on some properties of statistical methods largely unknown. For instance, our experiments show that a statistical weighting scheme based on the hypergeometric distribution such as lexical specificity \cite{lafon1980variabilite} can perform at least as well as or better than tf-idf \cite{jones1972statistical}, while having additional advantages with respect to flexibility and efficiency. As for the graph-based methods, they can be more reliable than statistical methods without being considerably slower in practice. In fact, graph-based methods
initialized with tf-idf or lexical specificity performs best overall.









\section{Keyword Extraction}\label{sec:keywordExtraction}

\begin{figure}[t]
  \centering
  \includegraphics[width=0.95\linewidth]{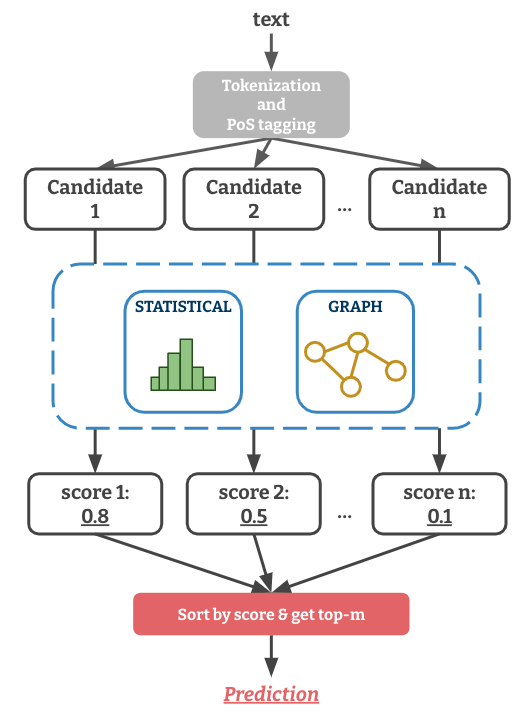}
  \caption{\label{fig:diagram} Overview of the keyword extraction pipeline.}
\end{figure}

Given a document with $m$ words $[w_1 \cdots w_{m}]$, keyword extraction is a task to find $n$ noun phrases, which can comprehensively represent the document.
As each of such phrases consists of contiguous words in the document, the task can be seen as an ordinary ranking problem over all candidate phrases appeared in the document.
A typical keyword extraction pipeline is thus implemented as, first, to construct a set of candidate phrases $\mathcal{P}_d$ for a target document $d$ and, second, to compute importance scores for all of individual words in $d$.\footnote{In the case of multi-token candidate phrases, this score is averaged among its tokens.} 
Finally, the top-$n$ phrases
$\{y_j | j=1\ldots n \} \subset \mathcal{P}_d$ in terms of the aggregated score are selected as the prediction \cite{mihalcea2004textrank}. Figure~\ref{fig:diagram} shows an overview of the overarching methodology for unsupervised keyword extraction.

To compute word-level scores, there are mainly two types of approach: statistical and graph-based. There are also contributions that focus on training supervised models for keyword extraction \cite{witten2005kea,liu2010supervised}. However, due to the absence of large labeled data and domain-specificity, most efforts are still unsupervised, which is the focus of this paper.

\subsection{Statistical Models}\label{sec:statisticalMethods}
A statistical model attains an importance score based on word-level statistics or surface features, such as the word frequency or the length of word.\footnote{The term \textit{Statistical} may not be strictly accurate to refer to tf-idf or purely frequency-based models, but in this case we follow previous conventions by grouping all these methods based on word-level frequency statistics as \textit{statistical} \cite{aizawa2003information}.}
A simple keyword extraction method could be to simply use term frequency (tf) as a scoring function for each word, which tend to work reasonably well. However, this simple measure may miss important information such as the relative importance of a given word in a corpus. For instance, prepositions such as \textit{in} or articles such as \textit{the} tend to be highly frequent in a text corpus. However, they barely represent a keyword in a given text document. To this end, different variants have been proposed, which we summarize in two main alternatives: tf-idf (Section \ref{tf-idf}) and {\it lexical specificity} (Section \ref{lexspec}).

\subsubsection{TF-IDF}
\label{tf-idf}

As an extension of tf, term frequency–inverse document frequency (tf-idf) \cite{jones1972statistical} is one of most popular and effective methods  used for statistical keyword extraction \cite{el2009kp}, as well as still being an important component in modern information retrieval applications \cite{marcos2020information,guu2020retrieval}.

Given a set of documents $\mathcal{D}$ and a word $w$ from a document $d\in\mathcal{D}$, tf-idf is defined as the proportion between its word frequency and its inverse document frequency\footnote{While there are other formulations and normalization techniques for tf-idf \cite{paik2013novel}, in this paper we focus on the traditional inverse-document frequency formulation.}, as
\begin{equation}
    \label{eq:tfidf}
    s_{\text{{\it tfidf}}}(w|d) = {\text{{\it tf}}}(w|d) \cdot \log_2\frac{| \mathcal{D} |}{{\text{{\it df}}}(w|\mathcal{D})}
\end{equation}
where we define $|\cdot|$ as the number of elements in a set,
\textit{tf}$(w|d)$ as a frequency of $w$ in $d$,
and \textit{df}$(w|\mathcal{D})$ as a document frequency of $w$ over a dataset $\mathcal{D}$.
In practice, \textit{tf}$(w|d)$ is often computed by counting the number of times that $w$ occurs in $d$, while \textit{df}$(w|\mathcal{D})$ by the number of documents in $\mathcal{D}$ that contain $w$. 

To give a few examples of statistical models based on tf-idf and its derivatives in a keyword extraction context, KP-miner \cite{el2009kp} utilizes tf-idf, a word length, and the absolute position of a word in a document to determine the importance score, while RAKE \cite{rose2010automatic} uses the term degree, the number of different word it co-occurs with, divided by tf. 
Recently, YAKE \cite{campos2020yake} established strong baselines on public datasets by combining various statistical features including casing, sentence position, term/sentence-frequency, and term-dispersion. In this paper, however, we focus on the vanilla implementation of term frequency and tf-idf.

\subsubsection{Lexical specificity}
\label{lexspec}


Lexical specificity \cite{lafon1980variabilite} is a statistical metric to extract relevant words from a subcorpus using a larger corpus as reference. In short, lexical specificity extracts a set of most representative words for a given text based on the hypergeometric distribution. The hypergeometric distribution represents the discrete probability of \textit{k} successes in \textit{n} draws, without replacement. In the case of lexical specificity, \textit{k} represents the word frequency and \textit{n} the size of a corpus. While not as widely adoped as tf-idf, lexical specificity has been used in similar term extraction tasks \citep{drouin2003term}, but also in textual data analysis \citep{Lebartetal:1998}, domain-based disambiguation \citep{billami2014annotation}, or as a weighting scheme for building vector representations for concepts and entities  \citep{camacho2016nasari} or sense embeddings \cite{scarlini2020sensembert} in NLP.

Formally, the lexical specificity for a word $w$ in a document $d$ is defined as
\begin{equation}
    \label{eq:lexical-specificity}
    s_{\text{{\it spec}}}(w|d) = - \log_{10} \sum_{l=f}^{F} P_{hg}(x{=}l, m_d, M, f, F)
\end{equation}
where $m_d$ is the total number of words in $d$ and $P_{hg}(x{=}l, m, M, f, F)$ represents the probability of a given word to appear $l$ times exactly in $d$ according to the hypergeometric distribution parameterised with $m_d$, $M$, $f$, and $F$, which are defined as below.
\begin{equation}
    M=\sum_{d\in \mathcal{D}}m_d,\: f={\text{{\it tf}}}(w|d), \: F= \sum_{d\in \mathcal{D}} {\text{{\it tf}}}(w|d)
\end{equation}

Note also that, unlike in tf-idf, for lexical specificity a perfect partition of documents of $\mathcal{D}$  (reference corpus) is not required. This also opens up to other possibilities, such as using larger corpora as reference, for example.  

\subsection{Graph-based Methods}\label{graphBasedeMethods}
The basic idea behind graph-based methods is to identify the most relevant words from a graph constructed from a text document, where words are nodes and their connections are measured in different ways \cite{beliga2015overview}. For this, PageRank \cite{page1999pagerank} and its derivatives have proved to be highly successful \cite{mihalcea2004textrank,wan2008collabrank,florescu2017positionrank,sterckx2015topical,bougouin2013topicrank}.


Formally, let $\mathcal{G}=(\mathcal{V}, \mathcal{E})$ be a graph where $\mathcal{V}$ and $\mathcal{E}$ are its associated set of vertices and edges. In a typical word graph construction on a document $d$ \cite{mihalcea2004textrank}, $\mathcal{V}$ is defined as the set of all unique words in $d$ and each edge $e_{w_i, w_j}\in\mathcal{E}$ represents a strength of the connection between two words $w_i,w_j\in \mathcal{V}$.
Then, a Markov chain from $w_j$ to $w_i$ on a word graph can be defined as
\begin{equation}
    \label{eq:random-surfer}
    p(w_i | w_j) = (1-\lambda) \frac{e_{w_i,w_j}}{\sum_{w_k \in \mathcal{V}_i} e_{w_i,w_k}} + \lambda p_{b}(w_i)
\end{equation}
where
$\mathcal{V}_i\subset \mathcal{V}$ is a set of incoming nodes to $w_i$,
$p_{b}(\cdot)$ is a prior probabilistic distribution over $\mathcal{V}$, and 
$0 \leq \lambda \leq 1$ is a parameter to control the effect of $p_{b}(\cdot)$. This probabilistic model \eqref{eq:random-surfer} is commonly known as the random surfer model \cite{page1999pagerank}. The prior term $p_{b}(\cdot)$, which is originally a uniform distribution, is introduced to enable any transitions even if there are no direct connections among them. 
Once a word graph is built, PageRank is applied to estimate a probability $\hat{p}(w)$ for every word $w\in \mathcal{V}$, which is used as an importance score. 

TextRank \cite{mihalcea2004textrank} uses an undirected graph and defines the edge weight as $e_{w_i, w_j} = 1$ if $w_i$ and $w_j$ co-occurred within $l$ contiguous sequence of words in $d$, otherwise $e_{w_i, w_j} = 0$.
SingleRank \cite{wan2008collabrank} extends TextRank by modifying the edge weight as the number of co-occurrence of $w_i$ and $w_j$ within the $l$-length sliding window and ExpandRank \cite{wan2008single} multiplies the weight by cosine similarity of tf-idf vector within neighbouring documents. To reflect a statistical prior knowledge to the estimation, recent works proposed to use non-uniform distributions for $p_{b}(\cdot)$. \citet{florescu2017positionrank} observed that keywords are likely to occur very close to the first few sentences in a document in academic paper and proposed PositionRank in which $p_{b}(\cdot)$ is defined as the inverse of the absolute position of each word in a document.
TopicalPageRank (TPR) \cite{jardine2014topical,sterckx2015topical} introduces a topic distribution inferred by Latent Dirichlet Allocation (LDA) as a $p_{b}(\cdot)$, so that the estimation contains more semantic diversity across topics.
TopicRank \cite{bougouin2013topicrank} clusters the candidates before running PageRank to group similar words together, and MultipartiteRank \cite{boudin-2018-unsupervised} extends it by employing a multipartite graph for a better candidate selection within a cluster.

\begin{table*}[t]
\centering
\small
\resizebox{\textwidth}{!}{
\begin{tabular}{ccccrrrrrrrrrrr}
\toprule
\multirow{3}{*}{Data} & \multirow{3}{*}{Size} & \multirow{3}{*}{Domain} & \multirow{3}{*}{Type} & \multirow{3}{*}{Divers.} & \multicolumn{2}{c}{\multirow{2}{*}{\# NPs}} & \multicolumn{2}{c}{\multirow{2}{*}{\# tokens}} & \multicolumn{2}{c}{\multirow{2}{*}{Vocab size}} & \multicolumn{4}{c}{\# keyphrases}                                              \\ \cline{12-15}
                      &                       &                         &                       &                            & \multicolumn{2}{r}{}                            & \multicolumn{2}{r}{}                          & \multicolumn{2}{r}{}                           & \multicolumn{2}{c}{total} & \multicolumn{2}{c}{multi-words} \\ \cline{6-15}
                      &                       &                         &                       &                            & avg                    & std                    & avg                   & std                   & avg                    & std                   & avg             & std            & avg                  & std                \\ \midrule
KPCrowd               & 500                   & -                       & news                     & 0.44                        & 77                   & 62.0                   & 447                 & 476.7                 & 197                  & 140.6                 & 16.5            & 12.0           & 3.7                  & 3.8                \\
Inspec                & 2000                  & CS                      & abstract                    & 0.55                        & 27                   & 12.3                   & 138                 & 66.6                  & 76                   & 28.6                  & 5.8             & 3.5            & 4.8                  & 3.2                \\
Krapivin2009          & 2304                  & CS                      & article                    & 0.12                        & 815                  & 252.1                  & 9131                & 2524.4                & 1081                 & 256.2                 & 3.8             & 2.1            & 2.9                  & 1.9                \\
Nguyen2007            & 209                   & -                       & article                     & 0.15                        & 618                  & 113.9                  & 5931                & 1023.1                & 909                  & 142.4                 & 7.2             & 4.3            & 4.8                  & 3.2                \\
PubMed                & 500                   & BM                      & article                    & 0.18                        & 566                  & 196.5                  & 4461                & 1626.4                & 800                  & 223.7                 & 5.7             & 2.7            & 1.5                  & 1.3                \\
Schutz2008            & 1231                  & BM                      & article                    & 0.29                        & 630                  & 287.7                  & 4201                & 2251.1                & 1217                 & 468.4                 & 28.5            & 10.3           & 10.1                 & 4.9                \\
SemEval2010           & 243                   & CS                      & article                    & 0.13                        & 898                  & 207.7                  & 9740                & 2443.4                & 1218                 & 209.1                 & 11.6            & 3.3            & 8.8                  & 3.3                \\
SemEval2017           & 493                   & -                       & paragraph                     & 0.54                        & 40                   & 12.9                   & 198                 & 60.3                  & 106                  & 27.5                  & 9.3             & 4.9            & 6.3                  & 3.4                \\
citeulike180          & 183                   & BI                      & article                    & 0.21                        & 822                  & 173.0                  & 5521                & 978.8                 & 1171                 & 202.9                 & 7.8             & 3.4            & 1.1                  & 1.0                \\
fao30                 & 30                    & AG                      & article                    & 0.21                        & 774                  & 93.2                   & 5438                & 927.5                 & 1125                 & 157.1                 & 15.9            & 5.6            & 5.5                  & 2.6                \\
fao780                & 779                   & AG                      & article                    & 0.19                        & 776                  & 147.2                  & 5591                & 902.4                 & 1087                 & 210.3                 & 4.2             & 2.3            & 1.6                  & 1.3                \\
theses100             & 100                   & -                       & article                     & 0.21                        & 728                  & 131.3                  & 5397                & 958.4                 & 1134                 & 192.3                 & 2.4             & 1.5            & 0.8                  & 0.8                \\
kdd                   & 755                   & CS                      & abstract                    & 0.59                       & 16                   & 17.0                   & 82                  & 93.0                  & 48                   & 45.7                  & 0.7             & 0.9            & 0.6                  & 0.8                \\
wiki20                & 20                    & CS                      & report                    & 0.15                        & 817                  & 322.4                  & 7146                & 3609.8                & 1088                 & 295.4                 & 12.8            & 3.2            & 6.7                  & 2.7                \\
www                   & 1330                  & CS                      & abstract                    & 0.58                        & 18                   & 16.5                   & 91                  & 89.1                  & 53.0                   & 43.3                  & 0.9             & 1.0            & 0.5                  & 0.7                \\ \bottomrule
\end{tabular}
}

\caption{
\label{tab:data}
Dataset statistics, where size refers to the number of documents; diversity refers to a measure of variety of vocabulary computed as the number of unique words divided by the total number of words; number of noun phrases (NPs) refers to candidate phrases extracted by our pipeline; number of tokens is the size of the dataset; vocab size is the number of unique tokens, and number of keyphrase shows the statistics of gold keyphrases for which we report the total number keyphrases, as well as the number of keyphrases composed by more than one token (multi-tokens). In terms of statistics, we show the average (avg) and the standard deviation (std).
}

\end{table*}

Finally, there are a few other works that directly run graph clustering \cite{liu2009clustering,grineva2009extracting}, using edges to connect clusters instead of words, with semantic relatedness as a weight. Although these techniques can capture high-level semantics, the relatedness-based weights rely on external resources such as Wikipedia \cite{grineva2009extracting}, and thus add another layer of complexity in terms of generalization. For these reasons, they are excluded from this study.

\section{Experimental Setting}\label{sec:experimental-setting}
In this section, we explain our keyword extraction experimental setting. All our experiments are run on a 16-core Ubuntu 
computer equipped with 3.8GHz i7 core and 64GiB memory.\footnote{All the details to reproduce our experiments are available at \url{https://github.com/asahi417/kex}}

\paragraph{Datasets.}\label{sec:datasets}
To evaluate the keyword extraction methods, we consider 15 different public datasets in English.\footnote{All the datasets were fetched from a public data repository for keyword extraction data: \url{https://github.com/LIAAD/KeywordExtractor-Datasets}: KPCrowd \cite{marujo2013keyphrase}, Inspec \cite{hulth-2003-improved}, Krapivin2009 \cite{krapivin2009large}, SemEval2017 \cite{augenstein-etal-2017-semeval}, kdd \cite{gollapalli2014extracting}, www \cite{gollapalli2014extracting}, wiki20 \cite{medelyan2008domain}, PubMed \cite{schutz2008keyphrase}, Schutz2008 \cite{schutz2008keyphrase}, citeulike180 \cite{medelyan-etal-2009-human}, fao30 and fao780 \cite{medelyan2008domain}, guyen2007 \cite{nguyen2007keyphrase}, and SemEval2010 \cite{kim-etal-2010-semeval}.} Each entry in a dataset consists of a source document and a set of gold keyphrases, where the source document is processed through the pipeline described in Section~\ref{sec:preprocess} and the gold keyphrase set is filtered to include only phrases which appear in its candidate set. Table~\ref{tab:data} provides high-level statistics of each dataset, including length and number of keyphrases\footnote{We use keyword and keyphrase almost indistinctly, as some datasets contain keyphrases of more than a single token.} (both average and standard deviation). 




\begin{table*}[!t]
{\small
\centering
\scalebox{0.95}{
\begin{tabular}{l|l|rrrr|rrrrrrr}
\toprule
\multirow{3}{*}{Metric} & \multirow{3}{*}{Dataset} & \multicolumn{4}{c}{Statistical} & \multicolumn{7}{c}{Graph-based} \\
& & \multirow{2}{*}{FirstN} & \multirow{2}{*}{TF} & Lex  & \multirow{2}{*}{TFIDF} & Text & Single & Position & Lex  & TFIDF & Single & Topic \\
& &                         &                     & Spec &                        & Rank & Rank   & Rank     & Rank & Rank  & TPR       & Rank \\ \midrule
\multirow{16}{*}{P@5} & KPCrowd      & 35.8          & 25.3 & \textbf{39.0} & \textbf{39.0} & 30.6          & 30.5          & 31.8          & 32.0          & 32.1          & 26.9          & 37.0          \\
                      & Inspec       & 31.0          & 18.9 & 31.0          & 31.5          & 33.2          & \textbf{33.8} & 32.7          & 32.9          & 33.3          & 30.4          & 31.3          \\
                      & Krapivin2009 & \textbf{16.7} & 0.1  & 8.7           & 7.6           & 6.6           & 9.1           & 14.3          & 9.7           & 9.7           & 7.4           & 8.5           \\
                      & Nguyen2007   & 17.8          & 0.2  & 17.2          & 15.9          & 13.1          & 17.3          & \textbf{20.6} & 18.6          & 18.6          & 14.0          & 13.3          \\
                      & PubMed       & 9.8           & 3.6  & 7.5           & 6.7           & 10.1          & \textbf{10.6} & 10.1          & 8.9           & 8.8           & 9.3           & 7.8           \\
                      & Schutz2008   & 16.9          & 1.6  & 39.0          & 38.9          & 34.0          & 36.5          & 18.3          & 38.9          & 39.4          & 14.5          & \textbf{46.6} \\
                      & SemEval2010  & 15.1          & 1.5  & 14.7          & 12.9          & 13.4          & 17.4          & \textbf{23.2} & 16.8          & 16.6          & 12.8          & 16.5          \\
                      & SemEval2017  & 30.1          & 17.0 & 45.7          & \textbf{47.2} & 41.5          & 43.0          & 40.5          & 46.0          & 46.4          & 34.3          & 36.5          \\
                      & citeulike180 & 6.6           & 9.5  & 18.0          & 15.2          & 23.0          & 23.9          & 20.3          & 23.2          & \textbf{24.4} & 23.7          & 16.7          \\
                      & fao30        & 17.3          & 16.0 & 24.0          & 20.7          & 26.0          & 30.0          & 24.0          & 29.3          & 29.3          & \textbf{32.7} & 24.7          \\
                      & fao780       & 9.3           & 3.2  & 11.7          & 10.5          & 12.4          & 14.3          & 13.2          & 13.2          & 13.1          & \textbf{14.5} & 12.0          \\
                      & kdd          & 11.7          & 7.0  & 11.2          & 11.6          & 10.6          & 11.5          & 11.9          & 11.6          & \textbf{11.9} & 9.4           & 10.7          \\
                      & theses100    & 5.6           & 0.9  & \textbf{10.7} & 9.4           & 6.6           & 7.8           & 9.3           & 10.6          & 9.1           & 8.3           & 8.1           \\
                      & wiki20       & 13.0          & 13.0 & 17.0          & 21.0          & 13.0          & 19.0          & 14.0          & 22.0          & \textbf{23.0} & 19.0          & 16.0          \\
                      & www          & 12.2          & 8.1  & 11.9          & 12.2          & 10.6          & 11.2          & \textbf{12.6} & 11.6          & 11.7          & 10.2          & 11.2          \\
                      & AVG          & 16.6          & 8.4  & 20.5          & 20.0          & 19.0          & 21.1          & 19.8          & 21.7          & \textbf{21.8} & 17.8          & 19.8          \\\midrule
\multirow{16}{*}{MRR} & KPCrowd      & 60.1          & 45.5 & \textbf{73.6} & 72.4          & 62.4          & 61.6          & 64.0          & 65.8          & 65.2          & 50.2          & 60.7          \\
                      & Inspec       & 57.3          & 33.0 & 52.4          & 52.8          & 51.4          & 52.4          & 57.1          & 53.3          & 53.7          & 50.5          & \textbf{57.8} \\
                      & Krapivin2009 & \textbf{36.1} & 1.3  & 22.9          & 21.0          & 18.1          & 22.2          & 31.4          & 23.6          & 23.8          & 19.1          & 21.8          \\
                      & Nguyen2007   & 43.0          & 2.8  & 38.1          & 41.2          & 30.8          & 34.6          & \textbf{43.2} & 36.4          & 37.9          & 29.8          & 33.7          \\
                      & PubMed       & 23.1          & 13.3 & 23.5          & 21.4          & \textbf{31.7} & 30.5          & 30.6          & 26.9          & 26.3          & 26.0          & 19.8          \\
                      & Schutz2008   & 24.6          & 8.6  & 76.6          & \textbf{76.7} & 68.9          & 70.9          & 38.5          & 75.5          & 76.3          & 33.7          & 67.3          \\
                      & SemEval2010  & \textbf{49.7} & 4.5  & 35.8          & 34.6          & 32.9          & 35.5          & 47.8          & 35.3          & 36.4          & 28.7          & 35.9          \\
                      & SemEval2017  & 52.0          & 32.7 & 68.6          & \textbf{68.7} & 61.4          & 63.5          & 62.4          & 67.3          & 67.2          & 54.3          & 63.7          \\
                      & citeulike180 & 20.9          & 23.6 & 55.5          & 47.7          & 58.2          & 62.6          & 51.0          & 63.0          & \textbf{65.7} & 62.5          & 40.3          \\
                      & fao30        & 31.1          & 38.3 & 61.8          & 49.1          & 60.2          & 70.0          & 48.6          & 66.1          & 67.0          & \textbf{74.6} & 50.6          \\
                      & fao780       & 17.0          & 8.5  & 39.0          & 35.9          & 36.1          & 38.6          & 35.9          & \textbf{39.5} & 38.9          & 38.4          & 31.6          \\
                      & kdd          & 26.1          & 13.0 & 27.0          & 27.8          & 24.5          & 26.5          & 28.1          & 27.9          & \textbf{28.8} & 18.3          & 26.2          \\
                      & theses100    & 15.1          & 3.1  & \textbf{32.5} & 31.6          & 23.2          & 26.3          & 24.9          & 31.6          & 31.1          & 26.1          & 26.9          \\
                      & wiki20       & 27.5          & 27.7 & \textbf{52.7} & 47.7          & 40.1          & 45.7          & 31.1          & 52.2          & 46.5          & 39.6          & 35.5          \\
                      & www          & 29.7          & 17.1 & 30.5          & \textbf{30.6} & 26.5          & 27.6          & 30.4          & 29.2          & 30.1          & 21.7          & 27.9          \\
                      & AVG          & 34.2          & 18.2 & 46.0          & 44.0          & 41.8          & 44.6          & 41.7          & 46.2          & \textbf{46.3} & 38.2          & 40.0         \\\bottomrule
\end{tabular}
}
}
\caption{Mean precision at top 5 (P@5) and mean reciprocal rank (MRR). The best score in each dataset is highlighted using a bold font. }
\label{tab:main-result}
\end{table*}

\paragraph{Preprocessing.}\label{sec:preprocess}
Before running keyword extraction on each dataset, we apply standard text preprocessing operations.
The documents are first tokenized into words by segtok\footnote{\url{https://pypi.org/project/segtok/}}, a python library for tokenization and sentence splitting.
Then, each word is stemmed to reduce it to its base form for comparison purpose by Porter Stemmer from NLTK \cite{bird2009natural}, a widely used python library for text processing.
Part-of-speech annotation is carried out using NLTK tagger. To select a candidate phrase set $\mathcal{P}_d$, following the literature \cite{wan2008single}, we consider contiguous nouns in the document $d$ that form a noun phrase satisfying the regular expression \textsc{(adjective)*(noun)+}.\footnote{While the vast majority of keywords in the considered datasets follow this structure, there are a few cases of different Part-of-Speech tags as keywords, or where this simple formulation can miss a correct candidate. Nonetheless, our experimental setting is focused on comparing keyword extraction measures, within the same preprocessing framework.}
We then filter the candidates with a stopword list taken from the official YAKE implementation\footnote{\url{https://github.com/LIAAD/yake}} \cite{campos2020yake}. Finally, for the statistical methods and the graph-based methods based on them (i.e., LexRank and TFIDFRank), we compute prior statistics including term frequency (tf), tf-idf, and LDA by Gensim \cite{rehurek_lrec} within each dataset.

\paragraph{Comparison Models.}\label{sec:comparisonSystem}


As \textbf{statistical} models, we include keyword extraction methods based on tf, tf-idf, and lexical specificity referred as TF, TFIDF, and LexSpec\footnote{For lexical specificity, we follow the implementation of \citet{camacho2016nasari}. } respectively.\footnote{As mentioned in Section \ref{sec:statisticalMethods}, we do not include YAKE \cite{campos2020yake} as our experiments are focused on analyzing single features on their own in a unified setting.
YAKE utilizes a unified preprocessing and a combination of various textual features, 
which are out of scope in this paper.} Each model uses its statistics as a score for the individual words and then aggregates them to score the candidate phrases (see Section \ref{sec:statisticalMethods}). 
We also add a heuristic baseline which takes the first $n$ phrases as its prediction (FirstN). 
As \textbf{graph-based} models, we compare five distinct methods: TextRank \cite{mihalcea2004textrank}, SingleRank \cite{wan2008collabrank}, PositionRank \cite{florescu2017positionrank}, SingleTPR \cite{sterckx2015topical}, and TopicRank \cite{bougouin2013topicrank}.
Additionally, we propose two extensions of SingleRank, which we call TFIDFRank and LexRank, where a word distribution computed by tf-idf or lexical specificity is used for $p_b(\cdot)$. 
As implementations of graph operations such as PageRank and word graph construction, we use NetworkX \cite{SciPyProceedings_11}, a graph analyzer in Python.

\section{Results}\label{sec:results}

In this section, we report our main experimental results comparing unsupervised keyword extraction methods. Table~\ref{tab:main-result} shows the results obtained by all comparison systems.\footnote{Results are reported according to standard metrics in keyword extraction and IR: precision at 5 (P@5) and mean reciprocal rank (MRR). The appendix includes details about these metrics and results for additional metrics.} 
The algorithms in each metric that achieve the best accuracy across datasets are TFIDFRank for P@5, 
and LexSpec and TFIDF for MRR.
In the averaged metrics over all datasets, lexical specificity and tf-idf based models (TFIDF, LexSpec, TFIDFRank, and LexRank) are shown to perform high in general. 
In particular, the hybrid models LexRank and TFIDFRank achieve the best accuracy on all the metrics, with LexSpec and TFIDF being competitive in MRR. Overall, despite their simplicity, both lexical specificity and tf-idf appear to be able to exploit effective features for keyword extraction from a variety of datasets, and perform robustly to domain shifts including document size, format, as well as the source domain. Moreover, TF gives a remarkably low accuracy on every metric and the huge gap between TF and TFIDF can be interpreted as the improvement given by the normalization provided by the inverse document frequency. However, as we discuss in Section \ref{sec:tdidfVSlexspec}, this IDF normalization relies on a corpus partition, which may not be available in all cases. 
On the other hand, a measure such as lexical specificity only needs overall term frequencies, which may have advantages in practical settings. 
In the following sections we perform a more in-depth analysis on these results and the global performance of each type model.


\noindent \textbf{Execution time.}
In terms of efficiency for each algorithm, we report the average process time over 100 independent trials on the Inspec dataset in Table~\ref{tab:time}, which also includes the time to compute each statistical prior over the dataset.
In general, none of the models perform very slowly. Not surprisingly, statistical models are faster than graph-based models due to the overhead introduced by the PageRank algorithm, although as a drawback they need to perform prior statistical computations for each dataset beforehand.

\begin{table}[t]
\centering
\small
\scalebox{1.05}{
\begin{tabular}{ccccc}
\hline
\multirow{2}{*}{Prior}  & \multirow{2}{*}{Model} & Time        & Time  & Time  \\
                        &              & prior                 & total & per doc \\\hline
\multirow{3}{*}{tf}& TF          & \multirow{3}{*}{10.2} & 11.5  & 0.0058         \\
                        & LexSpec      &                       & 12.1  & 0.0061         \\
                        & LexRank      &                       & 25.5  & 0.0128         \\\hline
\multirow{2}{*}{tf-idf} & TFIDF  & \multirow{2}{*}{10.3} & 22.4  & 0.0112         \\
                        & TFIDFRank    &                       & 26.5  & 0.0133         \\\hline
LDA                     & SingleTPR    & 16.2                  & 29.4  & 0.0147         \\\hline
\multirow{5}{*}{-}      & FirstN       & \multirow{5}{*}{-}    & 11.5  & 0.0058         \\
                        & TextRank     &                       & 14.9  & 0.0075         \\
                        & SingleRank   &                       & 15.0  & 0.0075         \\
                        & PositionRank &                       & 15.0  & 0.0075         \\
                        & TopicRank    &                       & 19.0  & 0.0095         \\ \hline
\end{tabular}
}
\caption{\label{tab:time} Average clock time (sec.) to process the Inspec dataset over 100 independent trials.}
\end{table}


\section{Analysis}
\label{sec:analysis}

Following the main results presented in the previous section, we perform an analysis on different aspects of the evaluation. In particular, we focus on the agreement among methods, 
overall performance (Section \ref{sec:meanPrecisionAnalysis}), and the features related to each dataset leading to each method's performace (Section \ref{sec:regressionAnalysis}). 

\subsection{Mean Precision Analysis}\label{sec:meanPrecisionAnalysis}

The objective of the following statistical analysis is to compare the overall performance of the keyword extraction methods in terms of their mean performance (i.e., P@5 and MRR). For this analysis, all 117,447 documents are considered individually.

Table \ref{tab:precisionSummary} illustrates the mean P@5 and MMR for each key extraction method. Across all the metrics, the best results are obtained by TFIDFRank. The differences between the models are tested for statistical significance using paired Wilcoxon rank sum tests.\footnote{For the sake of space, the full statistical significance analyses are presented in the appendix. However, these are also commented in our discussion (Section \ref{sec:discussion}).} 
A method is said to dominate another in terms of performance if it is non-worse in all the metrics and strictly better in at least one metric. Following this rule, it is possible to rank the methods according to their dominance order (i.e., the Pareto ranking): the top methods are those that are non-dominated, followed by those that are dominated only by methods of the first group, et cetera. The resulting ranking, which only considers statistically significant differences, is presented in the following: (1) TFIDFRank; (2) LexRank and LexSpec; (3) SingleRank and TFIDF; (4) PositionRank and TopicRank; (5) TextRank; (6) FirstN; (7) SingleTPR; (8) TF.

As can be observed in this ranking and in the results of Table \ref{tab:precisionSummary}, the best method is TFIDFRank, which dominates all the others. LexSpec slightly but consistently outperforms TFIDF, which is an interesting result on its own given the predominance in the use of TFIDF in the literature and in practical applications. We extend the discussion about the comparison of LexSpec and TFIDF in Section \ref{sec:tdidfVSlexspec}.

\begin{table}[t]
  \centering
  \includegraphics[width=0.82\linewidth]{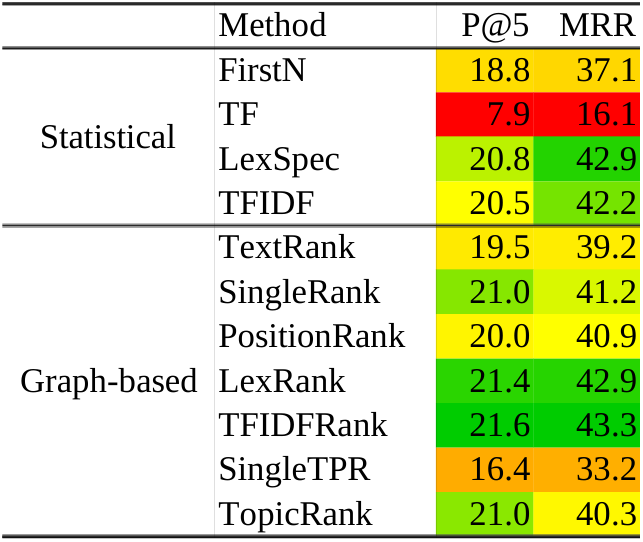}
  \caption{Key extraction methods' mean P@5 and MRR. Each column is independently colour-coded according to a gradient that goes from green (best/highest value) to red (worst/lowest value).}
  \label{tab:precisionSummary}
\end{table}

\begin{table*}[t]
\centering
\resizebox{\linewidth}{!}{
\begin{tabular}{rrc||cll}
\hline 
$m1$ & metric & \multicolumn{2}{c}{statistically significant variables} & metric & $m2$\tabularnewline
\hline 
\multirow{2}{*}{LexSpec} & P@5 & avg\_word ({*}{*}{*}), sd\_vocab ({*}) & sd\_word ({*}{*}{*}) & P@5 & \multirow{2}{*}{TFIDF}\tabularnewline
 & MRR & avg\_phrase ({*}{*}) avg\_keyword ({*}) & avg\_word ({*}{*}), avg\_vocab ({*}{*}) & MRR & \tabularnewline
\hline 
\multirow{2}{*}{SingleRank} & P@5 & avg\_phrase ({*}{*}), sd\_phrase ({*}{*}), avg\_keyword ({*}{*}) & sd\_word ({*}{*}), sd\_vocab ({*}{*}{*}), sd\_keyword ({*}) & P@5 & \multirow{2}{*}{TopicRank}\tabularnewline
 & MRR & avg\_phrase ({*}{*}{*}), avg\_keyword ({*}) & avg\_word ({*}{*}{*}), avg\_vocab ({*}{*}) & MRR & \tabularnewline
\hline 
\multirow{1}{*}{LexSpec} & P@5 & sd\_word ({*}), avg\_vocab ({*}{*}), sd\_keyword ({*}{*}) & avg\_phrase ({*}{*}), avg\_keyword ({*}{*}) & P@5 & \multirow{1}{*}{SingleRank}\tabularnewline
\hline 
\multirow{2}{*}{SingleRank} & P@5 & avg\_phrase ({*}), avg\_keyword ({*}) & sd\_word ({*}), avg\_vocab ({*}), sd\_keyword ({*}) & P@5 & \multirow{2}{*}{TFIDF}\tabularnewline
 & MRR & avg\_phrase ({*}), avg\_keyword ({*}) & sd\_word ({*}), avg\_vocab ({*}), sd\_keyword ({*}) & MRR & \tabularnewline
\hline 
\end{tabular}
}
\caption{Significant variables in the regression models comparing key-extraction methods' performance. Columns $m1$ and $m2$ report the compared methods; columns`metric' shows the performance metric considered; the central columns illustrate the statistically significant variables that positively affect the performance of each model. The significance of the variables is indicated between parenthesis, according to the following scale: 0 ‘***’ 0.001 ‘**’ 0.01 ‘*’ 0.05. Only models having $\mathrm{adj}R^2>0.5$ are included, as 
those are the variables that explain most of the differences in performance between the incumbent models.
\label{tab:significantVariables}}
\end{table*}

\subsection{Regression Analysis}\label{sec:regressionAnalysis}

The objective of this analysis is to understand what are a dataset's characteristics that make one method better than another at extracting keywords. For this purpose, a regression model is built for every performance metric (P@5 and MRR) and pair of key extraction methods (\emph{m1} and \emph{m2}). Formally, each observation is a pair in the Cartesian product $(\mathit{dataset}\times\mathit{method})$ in the regression models. The following independent variables are considered: avg\_word and sd\_word (i.e., average and standard deviation of the number of tokens in the dataset, representing the length of the documents); avg\_vocab and sd\_vocab (i.e., average and standard deviation of the number of unique tokens in the dataset, representing the lexical richness of the documents); avg\_phrase and sd\_phrase (i.e., average and standard deviation of the number of noun phrases in the dataset, representing the number of candidate keywords in the documents); avg\_keyword and sd\_keyword (i.e., average and standard deviation of the number of gold keyphrases associated to the dataset).\footnote{It can be noticed that not all the variables from Table \ref{tab:data} have been included in the regression analysis. The reasons for that are detailed in the following. The variable \textit{size} represents the number of documents in the dataset. As this is not a characteristic of the documents comprising the dataset, it has been disregarded (note that the size of each document is indeed included in the analysis, i.e., ‘avg\_word’). Variables \textit{domain} and \textit{type} are too sparse to be relevant. Finally, the variable \textit{diversity} is computed as $\frac{\mathrm{avg\_vocab}}{\mathrm{avg\_word}}$. Since both terms are already included in the regression model, adding diversity would result in an interdependence among the variables, consequently decreasing the interpretability of the results.} The regression models estimate the dependent variable as $\Delta\mathit{avg\_score} = \mathit{avg\_score}_{m1}-\mathit{avg\_score}_{m2}$, where $\mathit{avg\_score}_{m1}$ and $\mathit{avg\_score}_{m2}$ are the average performance metrics obtained by the methods \emph{m1} and \emph{m2} on the dataset's documents, respectively. Feature selection is carried out by forward stepwise-selection using BIC penalization to remove non-significant variables. Each model considers 15 observations and, overall, 110 regression models are fitted.

Given a regression model, its adjusted coefficient of determination ($\mathrm{adj}R^2$) is used as a measure of its goodness of fit. In fact, an $\mathrm{adj}R^2>0.50$ indicates that the independent variables explain most of the differences in performance between the models. The distribution of the $\mathrm{adj}R^2$ obtained by the regression models shows overall good explanatory capabilities: the 0\%, 25\%, 50\%, 75\%, and 100\% quantiles are 0,  0.6479,  0.7760, 0.8729, and  0.9776, respectively. 
Thus, $\sim$75\% of the models have an $\mathrm{adj}R^2>0.65$, and $\sim$50\% have $\mathrm{adj}R^2>0.78$, suggesting that, in general, the considered dataset's characteristics explain satisfactorily the differences in the results obtained by the key extraction methods. Therefore, the variables can be used to determine what method is more performant for a given dataset. In the rest of the paper, only the models having an $\mathrm{adj}R^2>0.50$ and their statistically significant variables (i.e., p-value < 0.05) are considered for interpretation.\footnote{More details about individual regression analyses and the significance of their variables are available in the appendix.}

The coefficients of the regression models can be used to understand under what circumstances each model is preferable. In fact, a positive coefficient identifies a variable that positively correlates with a greater precision for $m1$, while a negative coefficient corresponds to a variable that positively correlates with a greater precision for $m2$. Table \ref{tab:significantVariables} illustrates the significant variables for a selection of regression models. These are used in the following section to draw insights on the methods' preferences in terms of dataset features. 

\section{Discussion}\label{sec:discussion}

In this section, we provide a focused comparison among the different types of model, highlighting their main differences, advantages and disadvantages. 
First, we discuss the two main statistical methods analyzed in this paper, namely LexSpec and TFIDF (Section \ref{sec:tdidfVSlexspec}). Then, we analyze graphical methods, and in particular SingleRank and TopicRank (Section \ref{sec:singleVSTopic}). Finally, we provide an overview of the main differences between statistical and graph-based methods (Section \ref{sec:statisticalVSgraphical}).

\subsection{LexSpec vs. TFIDF} \label{sec:tdidfVSlexspec}
According to Table \ref{tab:precisionSummary}, LexSpec and TFIDF have similar average performance, although LexSpec obtains slightly better scores in both metrics. These differences are also statistically significant.
As for the Pareto ranking, LexSpec ranks second, while TFIDF ranks third. Therefore, the former should be preferred over the latter performance-wise.

However, TFIDF still performs better than LexSpec in certain datasets (see Table \ref{tab:main-result}). According to Table \ref{tab:significantVariables}, the choice of the key-extraction method strongly depends on the metric used. For P@5, TFIDF performs better in datasets having a higher variability in the number of words (sd\_word), while LexSpec prefers datasets with longer documents (avg\_word) and more variability in terms of lexical richness (sd\_vocab). For MRR, LexSpec exhibits a very different behaviour, performing significantly better in datasets with high average number of noun phrases (avg\_phrase) and high variability in the number of gold keywords (sd\_keyword). On the other hand, TFIDF prefers datasets with longer and lexically richer documents (avg\_word and avg\_vocab).

Broadly speaking, LexSpec and TFIDF have qualitative differences. Being based on the hypergeometric distribution, LexSpec has a statistical nature and probabilities can be directly inferred from it. 
While TFIDF can also be integrated within a probabilistic framework \cite{joachims1996probabilistic} or interpreted from an information-theoretic perspective \cite{aizawa2003information}, it is essentially heuristics-based. 
In practical terms, LexSpec has the advantage of not requiring a partition into documents unlike the traditional formulation of TFIDF. Moreover, given its statistical nature, LexSpec has been shown to be more robust to different document sizes \cite{camacho-collados-etal-2015-nasari}, as we could also empirically corroborate in the variable correlation analysis in the appendix. On the flip side, TFIDF is generally found to be relatively simple to tune for specific settings \cite{cui2014automatic}.

\subsection{SingleRank vs. TopicRank}
\label{sec:singleVSTopic}

In this analysis we compare two qualitatively different graph-based methods, namely Single Rank (a representative of vanilla graph-based methods) and TopicRank, which leverages topic models. The two methods have similar performances in term of P@5; however, SingleRank achieves a much better average MRR score, as illustrated in Table \ref{tab:precisionSummary}. The latter is also statistically significant. For this reason,  SingleRank completely dominates TopicRank. This is also reflected in the Pareto ranking, where the former ranks third and the latter fourth. Therefore, in general, SingleRank should be preferred.

The insights drawn from the regression models are summarised in the following. Table \ref{tab:significantVariables} shows that the performance of TopicRank depends on the metric used. On the other hand, SingleRank has a more stable set of preferences. However, it is still possible to identify a pattern. In fact, TopicRank is positively influenced by the number of words and the lexical richness of the documents in a dataset (sd\_word and sd\_vocab for P@5, and avg\_word and avg\_vocab for MRR), while SingleRank is affected by the number of noun phrases and keyphrases associated to the documents (avg\_phrase, sd\_phrase, and avg\_keyword).

\subsection{Statistical vs. Graph-based}
\label{sec:statisticalVSgraphical}
When comparing SingleRank versus TFIDF and LexSpec in terms of average performance (see Table \ref{tab:precisionSummary}), it can be seen that SingleRank performs better in terms of average P@5 (albeit only the difference with TFIDF is statistically significant); however, it performs worse than both the statistical methods in terms of average MRR (albeit no difference is statistically significant). Still, SingleRank ranks third (as TFIDF), because it is dominated by LexRank (all differences statistically significant). As LexSpec ranks second, it is recommendable to use this method instead of the other two. On the other hand, this is not a definitive argument in favour of using only statistical methods. In general, statistical methods tend to dominate in MRR over vanilla graph-based techniques. However, the method that achieves the highest scores across all the documents is TFIDFRank, which is graph-based in nature but initialized with TF-IDF. Results suggest that while statistical methods can be reliably used to extract relevant terms when precision is required (reminder that MRR rewards systems extracting the first correct candidate in top ranks), graphical methods can extract a more coherent set of keywords overall thanks to its graph-connectivity measures. This finding should be investigated more in detail in future research.

In terms of dataset features, Table \ref{tab:significantVariables} shows that the behaviour of SingleRank is very stable. In fact, across all metrics, SingleRank performs better for datasets with a high average of noun phrases and keyphrases (avg\_phrase and avg\_keyword). On the other hand, the statistical methods (i.e. TFIDF and LexSpec) achieve better performances on datasets with a high standard deviation for the number of words and keyphrases, and a high average number of unique tokens (sd\_word, sd\_keyword, and avg\_vocab). In conclusion, SingleRank performs better in datasets having a high number of candidate and gold keyphrases, while its performance is hindered in datasets having more lexical richness.

\paragraph{Efficiency and running time.} Statistical methods are shown to be faster overall in terms of computation time in our experiments (see Table \ref{tab:time}). However, all methods are overall efficient in practical settings, and this factor should not be of especial relevant unless computations need to be done on the fly or on a very large scale. As an advantage of graphical models, these do not require a prior computation over the whole dataset. Therefore, graph-based models could potentially reduce the gap in overall execution time in online learning settings, where new documents are added after the initial computations.

\section{Conclusion}
\label{sec:conclusion}

In this paper, we have presented a large-scale empirical comparison of unsupervised keyword extraction techniques. Our study was focused on two types of keyword extraction methods, namely statistical relying on frequency-based features, and graph-based exploiting the inter-connectivity of words in a corpus. Our analysis on fifteen diverse keyword extraction datasets revealed various insights with respect to each type of method. 

In addition to well-known term weighting schemes such as tf-idf, our comparison includes statistical methods such as lexical specificity, which shows better performance than tf-idf while being significantly less used in the literature. We have also explored various types of graph-based methods based on PageRank and on topic models, with varying conclusions with respect to performance and execution time. Our extensive evaluation and analysis can serve as a reference for future research to understand in detail the advantages and disadvantages of each approach in different settings, both qualitatively and quantitatively.

As future work, we plan to extend this analysis to fathom the extent and characteristics of the interactions of different methods and their complementarity. Moreover, we will extend this empirical comparison to other settings where the methods are used as weighting schemes for NLP and IR applications, and for languages other than English.

\section*{Acknowledgements}

Jose Camacho-Collados acknowledges support from the UKRI Future Leaders Fellowship scheme.

\bibliography{anthology,custom}

\begin{thebibliography}{54}
\expandafter\ifx\csname natexlab\endcsname\relax\def\natexlab#1{#1}\fi

\bibitem[{Aizawa(2003)}]{aizawa2003information}
Akiko Aizawa. 2003.
\newblock An information-theoretic perspective of tf--idf measures.
\newblock \emph{Information Processing \& Management}, 39(1):45--65.

\bibitem[{Arroyo-Fern{\'a}ndez et~al.(2019)Arroyo-Fern{\'a}ndez,
  M{\'e}ndez-Cruz, Sierra, Torres-Moreno, and Sidorov}]{arroyo2019unsupervised}
Ignacio Arroyo-Fern{\'a}ndez, Carlos-Francisco M{\'e}ndez-Cruz, Gerardo Sierra,
  Juan-Manuel Torres-Moreno, and Grigori Sidorov. 2019.
\newblock Unsupervised sentence representations as word information series:
  Revisiting tf--idf.
\newblock \emph{Computer Speech \& Language}, 56:107--129.

\bibitem[{Augenstein et~al.(2017)Augenstein, Das, Riedel, Vikraman, and
  McCallum}]{augenstein-etal-2017-semeval}
Isabelle Augenstein, Mrinal Das, Sebastian Riedel, Lakshmi Vikraman, and Andrew
  McCallum. 2017.
\newblock \href {https://doi.org/10.18653/v1/S17-2091} {{S}em{E}val 2017 task
  10: {S}cience{IE} - extracting keyphrases and relations from scientific
  publications}.
\newblock In \emph{Proceedings of the 11th International Workshop on Semantic
  Evaluation ({S}em{E}val-2017)}, pages 546--555, Vancouver, Canada.
  Association for Computational Linguistics.

\bibitem[{Beliga et~al.(2015)Beliga, Me{\v{s}}trovi{\'c}, and
  Martin{\v{c}}i{\'c}-Ip{\v{s}}i{\'c}}]{beliga2015overview}
Slobodan Beliga, Ana Me{\v{s}}trovi{\'c}, and Sanda
  Martin{\v{c}}i{\'c}-Ip{\v{s}}i{\'c}. 2015.
\newblock An overview of graph-based keyword extraction methods and approaches.
\newblock \emph{Journal of information and organizational sciences},
  39(1):1--20.

\bibitem[{Billami et~al.(2014)Billami, Camacho-Collados, Jacquey, and
  Kister}]{billami2014annotation}
Mokhtar-Boumeyden Billami, Jos{\'e} Camacho-Collados, Evelyne Jacquey, and
  Laurence Kister. 2014.
\newblock {Annotation s{\'e}mantique et validation terminologique en texte
  int{\'e}gral en SHS}.
\newblock In \emph{Proceedings of TALN}, pages 363--376.

\bibitem[{Bird et~al.(2009)Bird, Klein, and Loper}]{bird2009natural}
Steven Bird, Ewan Klein, and Edward Loper. 2009.
\newblock \emph{Natural language processing with Python: analyzing text with
  the natural language toolkit}.
\newblock " O'Reilly Media, Inc.".

\bibitem[{Boudin(2018)}]{boudin-2018-unsupervised}
Florian Boudin. 2018.
\newblock \href {https://doi.org/10.18653/v1/N18-2105} {Unsupervised keyphrase
  extraction with multipartite graphs}.
\newblock In \emph{Proceedings of the 2018 Conference of the North {A}merican
  Chapter of the Association for Computational Linguistics: Human Language
  Technologies, Volume 2 (Short Papers)}, pages 667--672, New Orleans,
  Louisiana. Association for Computational Linguistics.

\bibitem[{Bougouin et~al.(2013)Bougouin, Boudin, and
  Daille}]{bougouin2013topicrank}
Adrien Bougouin, Florian Boudin, and B{\'e}atrice Daille. 2013.
\newblock Topicrank: Graph-based topic ranking for keyphrase extraction.
\newblock In \emph{International joint conference on natural language
  processing (IJCNLP)}, pages 543--551.

\bibitem[{Camacho-Collados et~al.(2018)Camacho-Collados, Delli~Bovi,
  Espinosa-Anke, Oramas, Pasini, Santus, Shwartz, Navigli, and
  Saggion}]{camacho-collados-etal-2018-semeval}
Jose Camacho-Collados, Claudio Delli~Bovi, Luis Espinosa-Anke, Sergio Oramas,
  Tommaso Pasini, Enrico Santus, Vered Shwartz, Roberto Navigli, and Horacio
  Saggion. 2018.
\newblock \href {https://doi.org/10.18653/v1/S18-1115} {{S}em{E}val-2018 task
  9: Hypernym discovery}.
\newblock In \emph{Proceedings of The 12th International Workshop on Semantic
  Evaluation}, pages 712--724, New Orleans, Louisiana. Association for
  Computational Linguistics.

\bibitem[{Camacho-Collados et~al.(2015)Camacho-Collados, Pilehvar, and
  Navigli}]{camacho-collados-etal-2015-nasari}
Jos{\'e} Camacho-Collados, Mohammad~Taher Pilehvar, and Roberto Navigli. 2015.
\newblock \href {https://doi.org/10.3115/v1/N15-1059} {{NASARI}: a novel
  approach to a semantically-aware representation of items}.
\newblock In \emph{Proceedings of the 2015 Conference of the North {A}merican
  Chapter of the Association for Computational Linguistics: Human Language
  Technologies}, pages 567--577, Denver, Colorado. Association for
  Computational Linguistics.

\bibitem[{Camacho-Collados et~al.(2016)Camacho-Collados, Pilehvar, and
  Navigli}]{camacho2016nasari}
Jos{\'e} Camacho-Collados, Mohammad~Taher Pilehvar, and Roberto Navigli. 2016.
\newblock Nasari: Integrating explicit knowledge and corpus statistics for a
  multilingual representation of concepts and entities.
\newblock \emph{Artificial Intelligence}, 240:36--64.

\bibitem[{Campos et~al.(2020)Campos, Mangaravite, Pasquali, Jorge, Nunes, and
  Jatowt}]{campos2020yake}
Ricardo Campos, V{\'\i}tor Mangaravite, Arian Pasquali, Alipio Jorge, C{\'e}lia
  Nunes, and Adam Jatowt. 2020.
\newblock Yake! keyword extraction from single documents using multiple local
  features.
\newblock \emph{Information Sciences}, 509:257--289.

\bibitem[{Cui et~al.(2014)Cui, Mamou, Kingsbury, and
  Ramabhadran}]{cui2014automatic}
Jia Cui, Jonathan Mamou, Brian Kingsbury, and Bhuvana Ramabhadran. 2014.
\newblock Automatic keyword selection for keyword search development and
  tuning.
\newblock In \emph{2014 IEEE International Conference on Acoustics, Speech and
  Signal Processing (ICASSP)}, pages 7839--7843. IEEE.

\bibitem[{Drouin(2003)}]{drouin2003term}
Patrick Drouin. 2003.
\newblock Term extraction using non-technical corpora as a point of leverage.
\newblock \emph{Terminology}, 9(1):99--115.

\bibitem[{El-Beltagy and Rafea(2009)}]{el2009kp}
Samhaa~R El-Beltagy and Ahmed Rafea. 2009.
\newblock Kp-miner: A keyphrase extraction system for english and arabic
  documents.
\newblock \emph{Information systems}, 34(1):132--144.

\bibitem[{Florescu and Caragea(2017)}]{florescu2017positionrank}
Corina Florescu and Cornelia Caragea. 2017.
\newblock Positionrank: An unsupervised approach to keyphrase extraction from
  scholarly documents.
\newblock In \emph{Proceedings of the 55th Annual Meeting of the Association
  for Computational Linguistics (Volume 1: Long Papers)}, pages 1105--1115.

\bibitem[{Gollapalli and Caragea(2014)}]{gollapalli2014extracting}
Sujatha~Das Gollapalli and Cornelia Caragea. 2014.
\newblock Extracting keyphrases from research papers using citation networks.
\newblock In \emph{Proceedings of the AAAI Conference on Artificial
  Intelligence}, volume~28.

\bibitem[{Grineva et~al.(2009)Grineva, Grinev, and
  Lizorkin}]{grineva2009extracting}
Maria Grineva, Maxim Grinev, and Dmitry Lizorkin. 2009.
\newblock Extracting key terms from noisy and multitheme documents.
\newblock In \emph{Proceedings of the 18th international conference on World
  wide web}, pages 661--670.

\bibitem[{Guu et~al.(2020)Guu, Lee, Tung, Pasupat, and
  Chang}]{guu2020retrieval}
Kelvin Guu, Kenton Lee, Zora Tung, Panupong Pasupat, and Mingwei Chang. 2020.
\newblock Retrieval augmented language model pre-training.
\newblock In \emph{International Conference on Machine Learning}, pages
  3929--3938. PMLR.

\bibitem[{Hagberg et~al.(2008)Hagberg, Schult, and Swart}]{SciPyProceedings_11}
Aric~A. Hagberg, Daniel~A. Schult, and Pieter~J. Swart. 2008.
\newblock Exploring network structure, dynamics, and function using networkx.
\newblock In \emph{Proceedings of the 7th Python in Science Conference}, pages
  11 -- 15, Pasadena, CA USA.

\bibitem[{Hulth(2003)}]{hulth-2003-improved}
Anette Hulth. 2003.
\newblock \href {https://www.aclweb.org/anthology/W03-1028} {Improved automatic
  keyword extraction given more linguistic knowledge}.
\newblock In \emph{Proceedings of the 2003 Conference on Empirical Methods in
  Natural Language Processing}, pages 216--223.

\bibitem[{Jabri et~al.(2018)Jabri, Dahbi, Gadi, and Bassir}]{jabri2018ranking}
Siham Jabri, Azzeddine Dahbi, Taoufiq Gadi, and Abdelhak Bassir. 2018.
\newblock Ranking of text documents using tf-idf weighting and association
  rules mining.
\newblock In \emph{2018 4th International Conference on Optimization and
  Applications (ICOA)}, pages 1--6. IEEE.

\bibitem[{Jardine and Teufel(2014)}]{jardine2014topical}
James Jardine and Simone Teufel. 2014.
\newblock Topical pagerank: A model of scientific expertise for bibliographic
  search.
\newblock In \emph{Proceedings of the 14th Conference of the European Chapter
  of the Association for Computational Linguistics}, pages 501--510.

\bibitem[{Joachims(1996)}]{joachims1996probabilistic}
Thorsten Joachims. 1996.
\newblock A probabilistic analysis of the rocchio algorithm with tfidf for text
  categorization.
\newblock Technical report, Carnegie-mellon univ pittsburgh pa dept of computer
  science.

\bibitem[{Jones(1972)}]{jones1972statistical}
Karen~Sparck Jones. 1972.
\newblock A statistical interpretation of term specificity and its application
  in retrieval.
\newblock \emph{Journal of documentation}.

\bibitem[{Kim et~al.(2010)Kim, Medelyan, Kan, and
  Baldwin}]{kim-etal-2010-semeval}
Su~Nam Kim, Olena Medelyan, Min-Yen Kan, and Timothy Baldwin. 2010.
\newblock \href {https://www.aclweb.org/anthology/S10-1004} {{S}em{E}val-2010
  task 5 : Automatic keyphrase extraction from scientific articles}.
\newblock In \emph{Proceedings of the 5th International Workshop on Semantic
  Evaluation}, pages 21--26, Uppsala, Sweden. Association for Computational
  Linguistics.

\bibitem[{Krapivin et~al.(2009)Krapivin, Autaeu, and
  Marchese}]{krapivin2009large}
Mikalai Krapivin, Aliaksandr Autaeu, and Maurizio Marchese. 2009.
\newblock Large dataset for keyphrases extraction.

\bibitem[{Lafon(1980)}]{lafon1980variabilite}
Pierre Lafon. 1980.
\newblock Sur la variabilit{\'e} de la fr{\'e}quence des formes dans un corpus.
\newblock \emph{Mots. Les langages du politique}, 1(1):127--165.

\bibitem[{Lahiri et~al.(2017)Lahiri, Mihalcea, and Lai}]{lahiri2017keyword}
Shibamouli Lahiri, Rada Mihalcea, and Po-Hsiang Lai. 2017.
\newblock Keyword extraction from emails.
\newblock \emph{Nat. Lang. Eng.}, 23(2):295--317.

\bibitem[{Lebart et~al.(1998)Lebart, Salem, and Berry}]{Lebartetal:1998}
Ludovic Lebart, A~Salem, and Lisette Berry. 1998.
\newblock \emph{Exploring textual data}.
\newblock Kluwer Academic Publishers.

\bibitem[{Liu et~al.(2010)Liu, Liu, and Liu}]{liu2010supervised}
Fei Liu, Feifan Liu, and Yang Liu. 2010.
\newblock A supervised framework for keyword extraction from meeting
  transcripts.
\newblock \emph{IEEE Transactions on Audio, Speech, and Language Processing},
  19(3):538--548.

\bibitem[{Liu et~al.(2009)Liu, Li, Zheng, and Sun}]{liu2009clustering}
Zhiyuan Liu, Peng Li, Yabin Zheng, and Maosong Sun. 2009.
\newblock Clustering to find exemplar terms for keyphrase extraction.
\newblock In \emph{Proceedings of the 2009 conference on empirical methods in
  natural language processing}, pages 257--266.

\bibitem[{Marcos-Pablos and
  Garc{\'\i}a-Pe{\~n}alvo(2020)}]{marcos2020information}
Samuel Marcos-Pablos and Francisco~J Garc{\'\i}a-Pe{\~n}alvo. 2020.
\newblock Information retrieval methodology for aiding scientific database
  search.
\newblock \emph{Soft Computing}, 24(8):5551--5560.

\bibitem[{Marujo et~al.(2013)Marujo, Viveiros, and Neto}]{marujo2013keyphrase}
Luis Marujo, M{\'a}rcio Viveiros, and Jo{\~a}o Paulo da~Silva Neto. 2013.
\newblock Keyphrase cloud generation of broadcast news.
\newblock \emph{arXiv preprint arXiv:1306.4606}.

\bibitem[{Medelyan et~al.(2009)Medelyan, Frank, and
  Witten}]{medelyan-etal-2009-human}
Olena Medelyan, Eibe Frank, and Ian~H. Witten. 2009.
\newblock \href {https://www.aclweb.org/anthology/D09-1137} {Human-competitive
  tagging using automatic keyphrase extraction}.
\newblock In \emph{Proceedings of the 2009 Conference on Empirical Methods in
  Natural Language Processing}, pages 1318--1327, Singapore. Association for
  Computational Linguistics.

\bibitem[{Medelyan and Witten(2008)}]{medelyan2008domain}
Olena Medelyan and Ian~H Witten. 2008.
\newblock Domain-independent automatic keyphrase indexing with small training
  sets.
\newblock \emph{Journal of the American Society for Information Science and
  Technology}, 59(7):1026--1040.

\bibitem[{Mihalcea and Tarau(2004)}]{mihalcea2004textrank}
Rada Mihalcea and Paul Tarau. 2004.
\newblock Textrank: Bringing order into text.
\newblock In \emph{Proceedings of the 2004 conference on empirical methods in
  natural language processing}, pages 404--411.

\bibitem[{Nguyen and Kan(2007)}]{nguyen2007keyphrase}
Thuy~Dung Nguyen and Min-Yen Kan. 2007.
\newblock Keyphrase extraction in scientific publications.
\newblock In \emph{International conference on Asian digital libraries}, pages
  317--326. Springer.

\bibitem[{Page et~al.(1999)Page, Brin, Motwani, and
  Winograd}]{page1999pagerank}
Lawrence Page, Sergey Brin, Rajeev Motwani, and Terry Winograd. 1999.
\newblock The pagerank citation ranking: Bringing order to the web.
\newblock Technical report, Stanford InfoLab.

\bibitem[{Paik(2013)}]{paik2013novel}
Jiaul~H Paik. 2013.
\newblock A novel tf-idf weighting scheme for effective ranking.
\newblock In \emph{Proceedings of the 36th international ACM SIGIR conference
  on Research and development in information retrieval}, pages 343--352.

\bibitem[{Ramos et~al.(2003)}]{ramos2003using}
Juan Ramos et~al. 2003.
\newblock Using tf-idf to determine word relevance in document queries.
\newblock In \emph{Proceedings of the first instructional conference on machine
  learning}, volume 242, pages 29--48. Citeseer.

\bibitem[{{\v R}eh{\r u}{\v r}ek and Sojka(2010)}]{rehurek_lrec}
Radim {\v R}eh{\r u}{\v r}ek and Petr Sojka. 2010.
\newblock {Software Framework for Topic Modelling with Large Corpora}.
\newblock In \emph{{Proceedings of the LREC 2010 Workshop on New Challenges for
  NLP Frameworks}}, pages 45--50, Valletta, Malta. ELRA.

\bibitem[{Riedel et~al.(2017)Riedel, Augenstein, Spithourakis, and
  Riedel}]{riedel2017simple}
Benjamin Riedel, Isabelle Augenstein, Georgios~P Spithourakis, and Sebastian
  Riedel. 2017.
\newblock A simple but tough-to-beat baseline for the fake news challenge
  stance detection task.
\newblock \emph{arXiv preprint arXiv:1707.03264}.

\bibitem[{Rose et~al.(2010)Rose, Engel, Cramer, and Cowley}]{rose2010automatic}
Stuart Rose, Dave Engel, Nick Cramer, and Wendy Cowley. 2010.
\newblock Automatic keyword extraction from individual documents.
\newblock \emph{Text Mining: Applications and Theory}, pages 1--20.

\bibitem[{Scarlini et~al.(2020)Scarlini, Pasini, and
  Navigli}]{scarlini2020sensembert}
Bianca Scarlini, Tommaso Pasini, and Roberto Navigli. 2020.
\newblock Sensembert: Context-enhanced sense embeddings for multilingual word
  sense disambiguation.
\newblock In \emph{Proceedings of the AAAI Conference on Artificial
  Intelligence}, volume~34, pages 8758--8765.

\bibitem[{Schutz et~al.(2008)}]{schutz2008keyphrase}
Alexander~Thorsten Schutz et~al. 2008.
\newblock Keyphrase extraction from single documents in the open domain
  exploiting linguistic and statistical methods.

\bibitem[{Sterckx et~al.(2015)Sterckx, Demeester, Deleu, and
  Develder}]{sterckx2015topical}
Lucas Sterckx, Thomas Demeester, Johannes Deleu, and Chris Develder. 2015.
\newblock Topical word importance for fast keyphrase extraction.
\newblock In \emph{Proceedings of the 24th International Conference on World
  Wide Web}, pages 121--122.

\bibitem[{Sun et~al.(2020)Sun, Xiong, Liu, Liu, and Bao}]{sun2020joint}
Si~Sun, Chenyan Xiong, Zhenghao Liu, Zhiyuan Liu, and Jie Bao. 2020.
\newblock Joint keyphrase chunking and salience ranking with bert.
\newblock \emph{arXiv preprint arXiv:2004.13639}.

\bibitem[{Tang et~al.(2020)Tang, Li, and Li}]{tang2020improved}
Zhong Tang, Wenqiang Li, and Yan Li. 2020.
\newblock An improved term weighting scheme for text classification.
\newblock \emph{Concurrency and Computation: Practice and Experience},
  32(9):e5604.

\bibitem[{Wan and Xiao(2008{\natexlab{a}})}]{wan2008collabrank}
Xiaojun Wan and Jianguo Xiao. 2008{\natexlab{a}}.
\newblock Collabrank: towards a collaborative approach to single-document
  keyphrase extraction.
\newblock In \emph{Proceedings of the 22nd International Conference on
  Computational Linguistics (Coling 2008)}, pages 969--976.

\bibitem[{Wan and Xiao(2008{\natexlab{b}})}]{wan2008single}
Xiaojun Wan and Jianguo Xiao. 2008{\natexlab{b}}.
\newblock Single document keyphrase extraction using neighborhood knowledge.
\newblock In \emph{AAAI}, volume~8, pages 855--860.

\bibitem[{Witten et~al.(2005)Witten, Paynter, Frank, Gutwin, and
  Nevill-Manning}]{witten2005kea}
Ian~H Witten, Gordon~W Paynter, Eibe Frank, Carl Gutwin, and Craig~G
  Nevill-Manning. 2005.
\newblock Kea: Practical automated keyphrase extraction.
\newblock In \emph{Design and Usability of Digital Libraries: Case Studies in
  the Asia Pacific}, pages 129--152. IGI global.

\bibitem[{Wu et~al.(2008)Wu, Luk, Wong, and Kwok}]{wu2008interpreting}
Ho~Chung Wu, Robert Wing~Pong Luk, Kam~Fai Wong, and Kui~Lam Kwok. 2008.
\newblock Interpreting tf-idf term weights as making relevance decisions.
\newblock \emph{ACM Transactions on Information Systems (TOIS)}, 26(3):1--37.

\bibitem[{Xiong et~al.(2019)Xiong, Hu, Xiong, Campos, and
  Overwijk}]{xiong-etal-2019-open}
Lee Xiong, Chuan Hu, Chenyan Xiong, Daniel Campos, and Arnold Overwijk. 2019.
\newblock \href {https://doi.org/10.18653/v1/D19-1521} {Open domain web
  keyphrase extraction beyond language modeling}.
\newblock In \emph{Proceedings of the 2019 Conference on Empirical Methods in
  Natural Language Processing and the 9th International Joint Conference on
  Natural Language Processing (EMNLP-IJCNLP)}, pages 5175--5184, Hong Kong,
  China. Association for Computational Linguistics.

\end{thebibliography}
\bibliographystyle{acl_natbib}

\appendix


\begin{table*}[!t]
{\small
\centering
\begin{tabular}{l|l|rrrr|rrrrrrr}
\toprule
\multirow{3}{*}{Metric} & \multirow{3}{*}{Dataset} & \multicolumn{4}{c}{Statistical} & \multicolumn{7}{c}{Graph-based} \\
& & \multirow{2}{*}{FirstN} & \multirow{2}{*}{TF} & Lex  & \multirow{2}{*}{TFIDF} & Text & Single & Position & Lex  & TFIDF & Single & Topic \\
& &                         &                     & Spec &                        & Rank & Rank   & Rank     & Rank & Rank  & TPR       & Rank \\ \midrule
\multirow{16}{*}{P@10} & KPCrowd      & \textbf{33.1} & 23.7 & 31.8 & 32.0          & 24.6 & 25.0          & 26.9          & 26.8         & 27.0          & 23.5          & 32.6          \\
                       & Inspec       & 27.4          & 21.0 & 29.7 & 30.3          & 32.4 & \textbf{32.6} & 31.6          & 31.9         & 32.3          & 29.1          & 26.1          \\
                       & Krapivin2009 & \textbf{16.0} & 0.1  & 8.4  & 7.1           & 6.5  & 9.0           & 13.6          & 9.4          & 9.3           & 7.3           & 8.1           \\
                       & Nguyen2007   & 15.0          & 0.5  & 13.4 & 12.3          & 11.5 & 13.8          & \textbf{16.9} & 15.2         & 15.0          & 11.9          & 12.4          \\
                       & PubMed       & \textbf{8.1}  & 3.5  & 5.2  & 4.9           & 7.2  & 7.4           & 7.6           & 6.2          & 6.0           & 6.6           & 6.9           \\
                       & Schutz2008   & 14.4          & 4.2  & 28.3 & 28.0          & 25.5 & 27.3          & 17.2          & 28.4         & 28.8          & 13.2          & \textbf{41.1} \\
                       & SemEval2010  & 12.0          & 0.9  & 11.0 & 10.5          & 11.0 & 13.8          & \textbf{17.3} & 13.0         & 13.5          & 10.9          & 14.4          \\
                       & SemEval2017  & 28.3          & 20.2 & 40.7 & \textbf{41.5} & 39.1 & 39.6          & 38.2          & 41.0         & 41.3          & 34.0          & 29.9          \\
                       & citeulike180 & 5.6           & 7.7  & 11.8 & 10.4          & 14.4 & \textbf{15.9} & 14.8          & 14.8         & 15.8          & 15.7          & 12.2          \\
                       & fao30        & 15.3          & 12.7 & 15.3 & 13.7          & 20.3 & 22.3          & 19.0          & 20.3         & 21.0          & \textbf{23.7} & 18.0          \\
                       & fao780       & 7.5           & 3.2  & 9.3  & 8.1           & 10.2 & 11.4          & 11.1          & 10.8         & 10.7          & \textbf{11.8} & 9.9           \\
                       & kdd          & 11.2          & 6.7  & 10.9 & 11.4          & 10.2 & 11.0          & \textbf{11.5} & 11.2         & 11.5          & 9.0           & 10.4          \\
                       & theses100    & 4.6           & 0.7  & 8.8  & 7.3           & 5.5  & 6.7           & 8.5           & \textbf{9.3} & 8.0           & 7.1           & 6.5           \\
                       & wiki20       & 13.0          & 9.5  & 12.0 & 12.0          & 12.0 & 15.0          & 12.5          & 14.0         & 15.0          & \textbf{16.5} & 16.0          \\
                       & www          & 11.3          & 7.5  & 11.1 & 11.4          & 9.9  & 10.5          & \textbf{11.8} & 10.9         & 10.9          & 9.6           & 10.3          \\
                       & AVG          & 14.9          & 8.1  & 16.5 & 16.1          & 16.0 & 17.4          & 17.2          & 17.5         & \textbf{17.7} & 15.3          & 17.0    \\\bottomrule     
\end{tabular}
}
\caption{Mean precision at top 10 (P@10). The best score in each dataset is highlighted using a bold font.}
\label{tab:main-result-sup}
\end{table*}

\section{Graph-based Models Formula}\label{sec:graphForm}
Supposing that we have computed tf-idf for a given dataset $\mathcal{D}$, the prior distribution for TFIDFRank is defined as 
\begin{equation}
    p_b(w) = \frac{s_{\text{{\it tfidf}}}(w|d)}{\sum_{\tilde{w} \in \mathcal{V}} s_{\text{{\it tfidf}}}(\tilde{w}|d)}
\end{equation}
for $w\in\mathcal{V}$.
Likewise, LexRank relies on the pre-computed {\it lexical specificity} prior (see Section 2.1.2 of the main paper), which defines the prior distribution for $d$ as  
\begin{equation}
    p_b(w) = \frac{s_{\text{{\it spec}}}(w)}{\sum_{\tilde{w} \in \mathcal{V}} s_{\text{{\it spec}}}(\tilde{w})}.
\end{equation}
All remaining specifications follow SingleRank's graph construction procedure.


\section{Evaluation Metrics}\label{sec:evaluationMetrics}
To evaluate the keyword extraction models, we employ standard metrics in the literature in keyword extraction and information retrieval: precision at k (P@k) and mean reciprocal rank (MRR).
In general, precision at k is computed as
\begin{equation}
\label{eq:preck}
    \text{P@k} = \sum_{d=1}^{|\mathcal{D}|}\frac{|y_d \cap \hat{y}^{k}_d |}{ \min\{|y_d|, k \}}
\end{equation}
where $y_d$ is the set of gold keyphrases provided with a document $d$ in the dataset $\mathcal{D}$ and $\hat{y}^{k}_d$ is a set of estimated top-$k$ keyphrases from a model for the document. The minimum operation between the number of gold keyphrases and gold labels in the denominator of Eq. \ref{eq:preck} is included as to provide a measure between 0 and 1, given the varying number of gold labels. This formulation follows previous retrieval tasks with similar settings such as SemEval 2018 \cite{camacho-collados-etal-2018-semeval}.

MRR measures the ranking quality given by a model as follows:
\begin{align}
    \text{MRR} &=\frac{1}{|\mathcal{D}|} \sum_{d=1}^{|\mathcal{D}|}\frac{1}{
    \min\left\{k \:\bigl\vert\: \vert \hat{y}^k_d \cap y_d \vert \geq 1  \right\}
    }
\end{align}
In this case, MRR takes into account the position of the first correct keyword from the ranked list of predictions.
\section{Additional Results (P@10)}\label{sec:additionalRes}
In addition to the metrics used in the main paper (i.e., P@5 and MRR), in Table~\ref{tab:main-result-sup} we show the main results for precision at 10 (P@10).

\section{Agreement Analysis}\label{sec:agreement}
For a visualization purpose, we compute agreement scores over all possible pairs of models as the percentage of predicted keywords the two models have in common in the top-5 prediction, as displayed in Table~\ref{fig:agreement}. Interestingly, the most similar models in terms of the agreement score are TFIDFRank and LexRank. Not surprisingly, TFIDF and LexSpec also hold a very high similarity that implies those two statistical measures capture quite close features. However, they also have a few marked differences. Moreover, we can see that graph-based models provide fairly high agreement scores, except for TopicRank, which can be due to the difference in the word graph construction procedure. In fact, TopicRank unifies similar word before building a word graph and that results in such a distinct behaviour among graph-based models. In the discussion section, we investigate the relation among each model in more detail.

\begin{table}[!t]
  \centering
  \includegraphics[width=\linewidth]{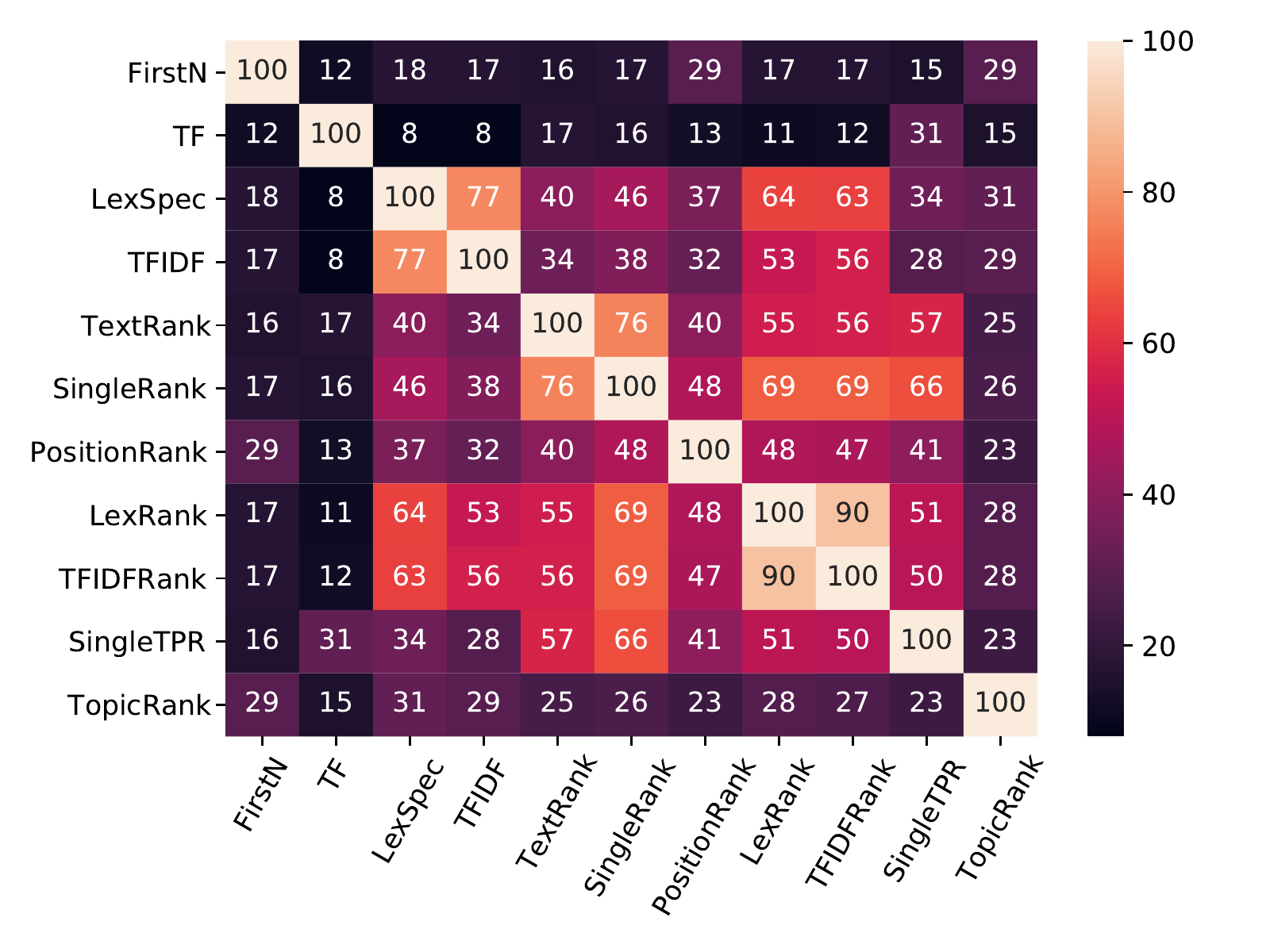}
  \caption{Overall pairwise agreement scores for the top 5 predictions.}
  \label{fig:agreement}
\end{table}

\section{Correlation Analysis}
\label{sec:correlation}

\begin{table}[!t]
  \centering
  \includegraphics[width=\linewidth]{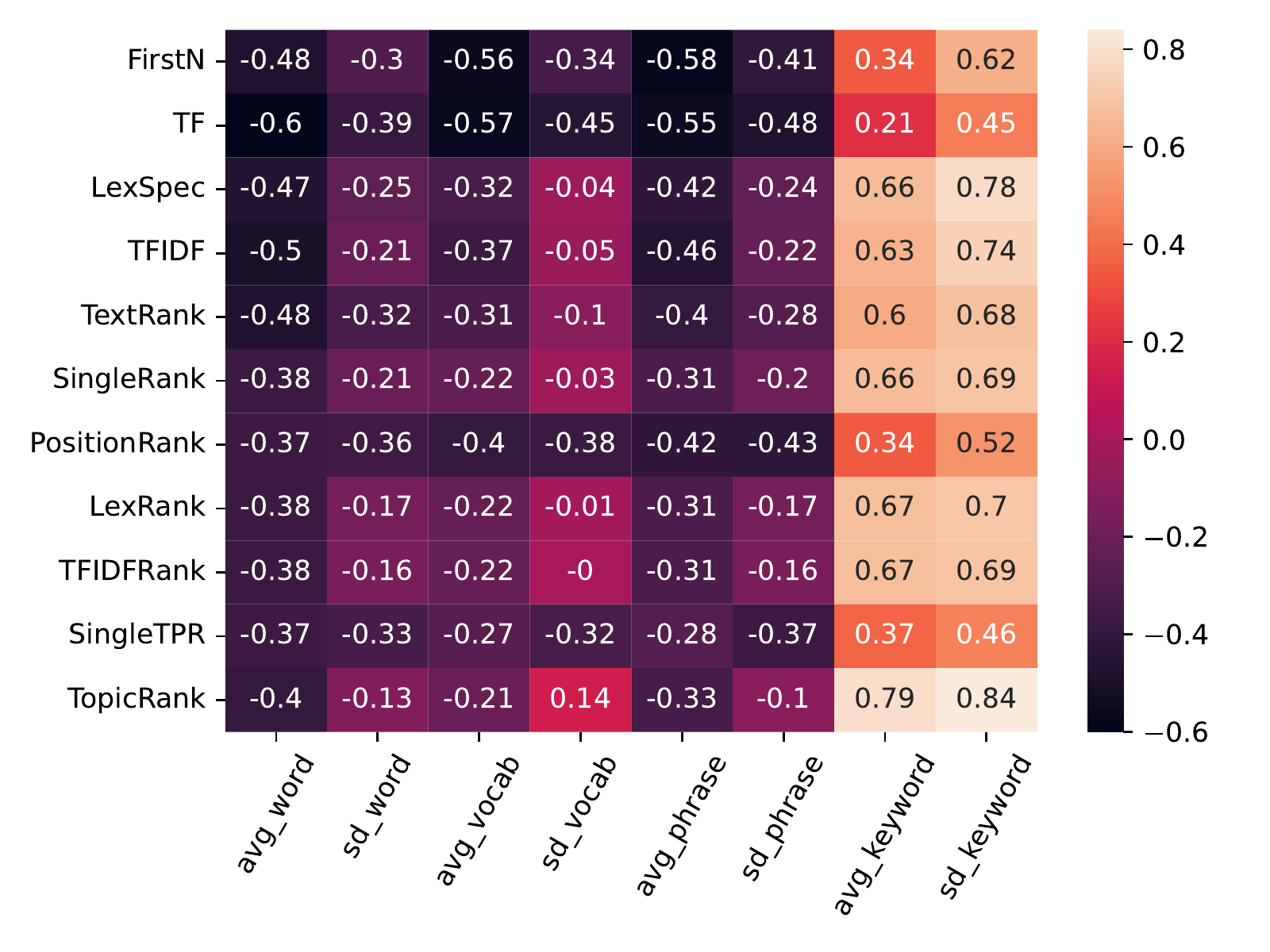}
  \caption{Correlations between algorithms and regression variables for metric P@5.}
  \label{fig:heatmap_pre5}
\end{table}

\begin{table}[!t]
  \centering
  \includegraphics[width=\linewidth]{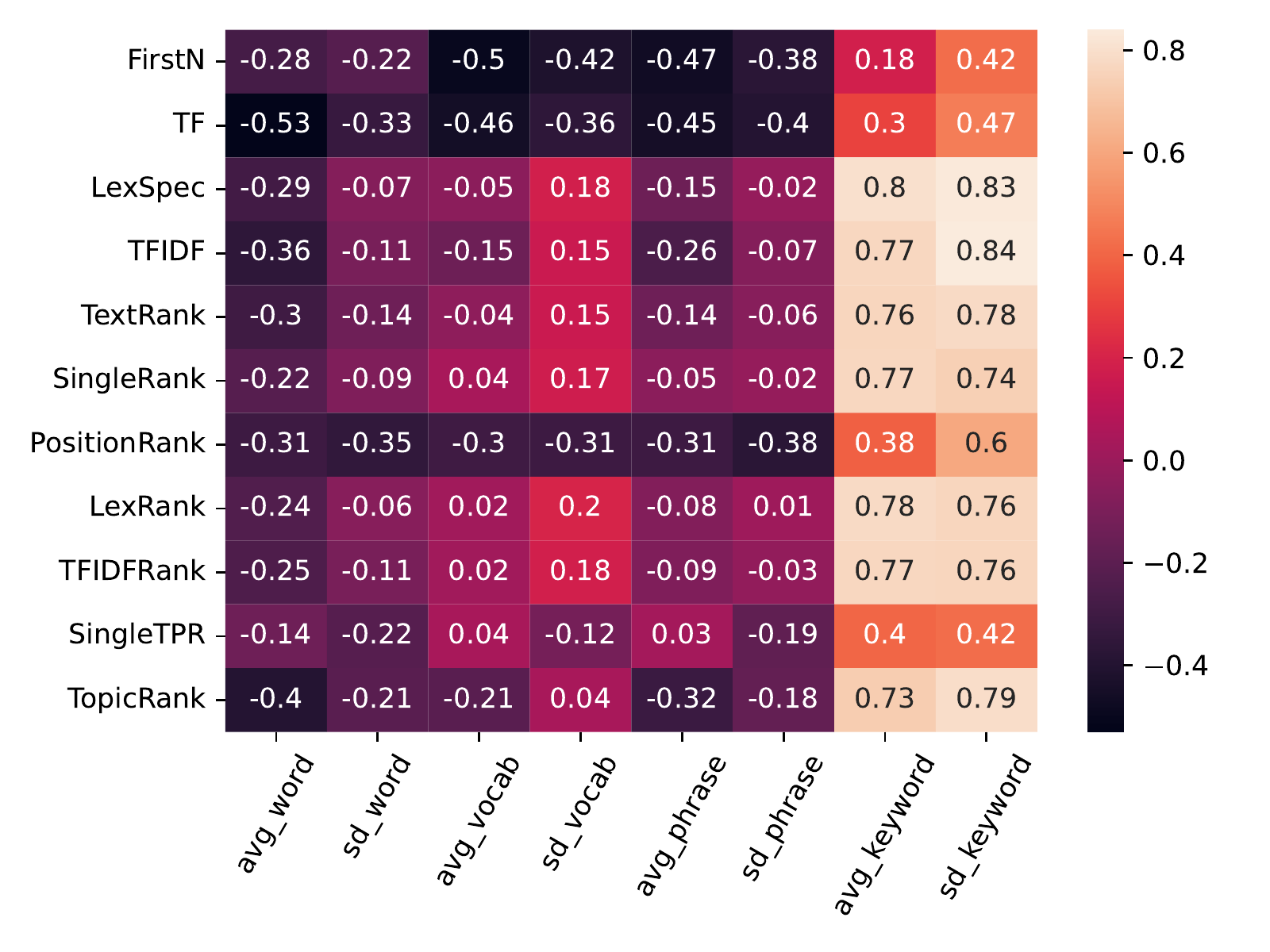}
  \caption{Correlations between algorithms and regression variables for metric MRR.}
  \label{fig:heatmap_mrr}
\end{table}

Tables \ref{fig:heatmap_pre5} and \ref{fig:heatmap_mrr} illustrate the correlation between algorithms and regression variables for P@5 and MMR, respectively. The results of this analysis can be interpreted in the following way: a model with a correlation close to zero can be said to be more robust towards the corresponding variable (i.e., less affected by changes in the variable's value) than another model with a higher absolute correlation. Overall, the scores of all the algorithms are higher for datasets with a high average number of gold keyphrases, while they are lower for datasets with a higher average number of tokens, unique tokens (thus, lexical richness), and candidate keywords. The standard deviations follow the same behavior as the averages.

\section{Statistical Significance}\label{sec:statisticalsign}
The performances of the algorithms on both P@5 and MRR have been tested to verify if they are statistically significant. As the data is not normally distributed ($\mathit{p-values}\sim0$ in Anderson-Darling normality tests), paired Wilcoxon rank sum tests have been used. The results are illustrated in Table \ref{tab:Wilcoxon}.

\begin{table*}[ht]
\centering
\resizebox{\textwidth}{!}{
\begin{tabular}{rrrrrrrrrrrr}
  \hline
 & FirstN & LexSpec & TF & TFIDF & TFIDFRank & TextRank & SingleRank & PositionRank & LexRank & SingleTPR & TopicRank \\ 
  \hline
FirstN & - & 0.00 & 0.00 & 0.00 & 0.00 & 0.07 & 0.00 & 0.00 & 0.00 & 0.00 & 0.00 \\ 
  LexSpec & 0.00 & - & 0.00 & 0.00 & 0.00 & 0.00 & 0.24 & 0.03 & 0.00 & 0.00 & 0.61 \\ 
  TF & 0.00 & 0.00 & - & 0.00 & 0.00 & 0.00 & 0.00 & 0.00 & 0.00 & 0.00 & 0.00 \\ 
  TFIDF & 0.00 & 0.00 & 0.00 & - & 0.00 & 0.00 & 0.00 & 0.69 & 0.00 & 0.00 & 0.02 \\ 
  TFIDFRank & 0.00 & 0.00 & 0.00 & 0.00 & - & 0.00 & 0.00 & 0.00 & 0.07 & 0.00 & 0.00 \\ 
  TextRank & 0.00 & 0.00 & 0.00 & 0.00 & 0.00 & - & 0.00 & 0.00 & 0.00 & 0.00 & 0.00 \\ 
  SingleRank & 0.00 & 0.27 & 0.00 & 0.40 & 0.00 & 0.00 & - & 0.00 & 0.00 & 0.00 & 0.37 \\ 
  PositionRank & 0.00 & 0.10 & 0.00 & 0.98 & 0.00 & 0.00 & 0.10 & - & 0.00 & 0.00 & 0.06 \\ 
  LexRank & 0.00 & 0.00 & 0.00 & 0.00 & 0.00 & 0.00 & 0.00 & 0.00 & - & 0.00 & 0.01 \\ 
  SingleTPR & 0.00 & 0.00 & 0.00 & 0.00 & 0.00 & 0.00 & 0.00 & 0.00 & 0.00 & - & 0.00 \\ 
  TopicRank & 0.00 & 0.00 & 0.00 & 0.00 & 0.00 & 0.00 & 0.03 & 0.08 & 0.00 & 0.00 & - \\ 
   \hline
\end{tabular}}
\caption{P-values of paired Wilcoxon rank sum tests on the P@5 score (upper triangular matrix) and MMR (lower triangular matrix) obtained by the algorithms on all the documents considered. A value below 0.05 indicates that the difference between the algorithms is statistically significant.}\label{tab:Wilcoxon}
\end{table*}

\section{Regression Models}
\label{sec:regressionmodels}

This section provides additional information regarding the regression models. In the following Tables \ref{tab:reg1} - \ref{tab:reg7}, the regression models for the comparisons considered in Section ``Discussion'' are presented. For each variable, the tables show the estimated coefficient value, the standard error, the t-value, and the p-value. The last column identifies the significance of the coefficient, according to the following scale: 0 ‘***’ 0.001 ‘**’ 0.01 ‘*’ 0.05 ‘.’ 0.1 ‘ ’ 1. The adjusted coefficient of determination ($\mathrm{adj}R^2$) is provided in the caption. Note that only the models having $\mathrm{adj}R^2>0.5$ are reported.


\begin{table}[h]
{\scriptsize
\begin{tabular}{lrrrrl}
\hline 
 & Estimate & Std. Error & t value & Pr(>|t|) & \tabularnewline
\hline 
(Intercept) & -0.7718 & 0.4430 & -1.7424 & 0.1093 & \\ 
  avg\_word & 0.0007 & 0.0001 & 6.2946 & 0.0001 & {*}{*}{*}\\ 
  sd\_word & -0.0029 & 0.0005 & -6.2922 & 0.0001 & {*}{*}{*}\\ 
  sd\_vocab & 0.0093 & 0.0032 & 2.8805 & 0.0150 & {*}\\ 
\hline 
\end{tabular}
}
\caption{LexSpec VS TFIDF; metric: P@5; $\mathrm{adj}R^2 = 0.76$.}\label{tab:reg1}
\end{table}

\begin{table}[h]
{\scriptsize
\begin{tabular}{lrrrrl}
\hline 
 & Estimate & Std. Error & t value & Pr(>|t|) & \tabularnewline
\hline 
(Intercept) & -0.9170 & 1.2298 & -0.7457 & 0.4730 \\ 
  avg\_word & -0.0027 & 0.0007 & -3.9583 & 0.0027 & {*}{*} \\ 
  avg\_vocab & -0.0383 & 0.0106 & -3.6158 & 0.0047 & {*}{*} \\ 
  avg\_phrase & 0.0800 & 0.0179 & 4.4689 & 0.0012 & {*}{*} \\ 
  avg\_keyword & 0.3064 & 0.1213 & 2.5251 & 0.0301 & {*} \\ 
\hline 
\end{tabular}
}
\caption{LexSpec VS TFIDF; metric: MRR; $\mathrm{adj}R^2 = 0.64$.}\label{tab:reg2}
\end{table}

\begin{table}[h]
{\scriptsize
\begin{tabular}{lrrrrl}
\hline 
 & Estimate & Std. Error & t value & Pr(>|t|) & \tabularnewline
\hline 
(Intercept) & 4.5423 & 1.0236 & 4.4377 & 0.0022 & {*}{*} \\ 
  sd\_word & -0.0096 & 0.0021 & -4.4760 & 0.0021 & {*}{*} \\ 
  sd\_vocab & -0.1277 & 0.0187 & -6.8144 & 0.0001 & {*}{*}{*} \\ 
  avg\_phrase & 0.0072 & 0.0021 & 3.4006 & 0.0094 & {*}{*} \\ 
  sd\_phrase & 0.1810 & 0.0375 & 4.8281 & 0.0013 & {*}{*} \\ 
  avg\_keyword & 0.6920 & 0.1726 & 4.0101 & 0.0039 & {*}{*} \\ 
  sd\_keyword & -0.9239 & 0.3274 & -2.8218 & 0.0224 & {*} \\ 
\hline 
\end{tabular}
}
\caption{SingleRank VS TopicRank; metric: P@5; $\mathrm{adj}R^2 = 0.88$.}\label{tab:reg3}
\end{table}

\begin{table}[h]
{\scriptsize
\begin{tabular}{lrrrrl}
\hline 
 & Estimate & Std. Error & t value & Pr(>|t|) & \tabularnewline
\hline 
(Intercept) & -1.8764 & 2.0988 & -0.8940 & 0.3923 \\ 
  avg\_word & -0.0061 & 0.0012 & -5.2350 & 0.0004 & {*}{*}{*} \\ 
  avg\_vocab & -0.0638 & 0.0181 & -3.5287 & 0.0055 & {*}{*} \\ 
  avg\_phrase & 0.1503 & 0.0306 & 4.9186 & 0.0006 & {*}{*}{*} \\ 
  avg\_keyword & 0.4663 & 0.2071 & 2.2518 & 0.0480 & {*} \\ 
\hline 
\end{tabular}
}
\caption{SingleRank VS TopicRank; metric: MRR; $\mathrm{adj}R^2 = 0.74$.}\label{tab:reg4}
\end{table}

\begin{table}[h]
{\scriptsize
\begin{tabular}{lrrrrl}
\hline 
 & Estimate & Std. Error & t value & Pr(>|t|) & \tabularnewline
\hline 
(Intercept) & -1.9469 & 1.1930 & -1.6320 & 0.1413 \\ 
  sd\_word & 0.0073 & 0.0023 & 3.1395 & 0.0138 & {*} \\ 
  avg\_vocab & 0.0397 & 0.0107 & 3.6991 & 0.0061 & {*}{*} \\ 
  avg\_phrase & -0.0575 & 0.0141 & -4.0726 & 0.0036 & {*}{*}\\ 
  sd\_phrase & -0.0518 & 0.0253 & -2.0479 & 0.0748 \\ 
  avg\_keyword & -1.0153 & 0.2214 & -4.5867 & 0.0018 & {*}{*} \\ 
  sd\_keyword & 1.9318 & 0.3985 & 4.8471 & 0.0013 & {*}{*} \\
\hline 
\end{tabular}
}
\caption{LexSpec VS SingleRank; metric: P@5; $\mathrm{adj}R^2 = 0.76$.}\label{tab:reg5}
\end{table}

\begin{table}[h]
{\scriptsize
\begin{tabular}{lrrrrl}
\hline 
 & Estimate & Std. Error & t value & Pr(>|t|) & \tabularnewline
\hline 
(Intercept) & 0.9891 & 1.7134 & 0.5773 & 0.5796 \\ 
  sd\_word & -0.0088 & 0.0034 & -2.6226 & 0.0305 & {*} \\ 
  avg\_vocab & -0.0388 & 0.0154 & -2.5141 & 0.0361 & {*} \\ 
  avg\_phrase & 0.0628 & 0.0203 & 3.0966 & 0.0147 & {*} \\ 
  sd\_phrase & 0.0476 & 0.0363 & 1.3103 & 0.2265 \\ 
  avg\_keyword & 0.9859 & 0.3179 & 3.1009 & 0.0146 & {*} \\ 
  sd\_keyword & -1.7774 & 0.5724 & -3.1051 & 0.0146 & {*} \\
\hline 
\end{tabular}
}
\caption{SingleRank VS TFIDF; metric: P@5; $\mathrm{adj}R^2 = 0.68$.}\label{tab:reg6}
\end{table}

\begin{table}[h]
{\scriptsize
\begin{tabular}{lrrrrl}
\hline 
 & Estimate & Std. Error & t value & Pr(>|t|) & \tabularnewline
\hline 
(Intercept) & 0.1791 & 3.6510 & 0.0491 & 0.9621 \\ 
  sd\_word & -0.0164 & 0.0049 & -3.3190 & 0.0106 & {*} \\ 
  avg\_vocab & -0.1326 & 0.0462 & -2.8690 & 0.0209 & {*} \\ 
  sd\_vocab & 0.1004 & 0.0544 & 1.8446 & 0.1023 \\ 
  avg\_phrase & 0.1921 & 0.0600 & 3.2029 & 0.0126 & {*} \\ 
  avg\_keyword & 2.1376 & 0.6667 & 3.2062 & 0.0125 & {*} \\ 
  sd\_keyword & -3.4630 & 1.2615 & -2.7451 & 0.0253 & {*} \\
\hline 
\end{tabular}
}
\caption{SingleRank VS TFIDF; metric: MRR; $\mathrm{adj}R^2 = 0.56$.}\label{tab:reg7}
\end{table}

\end{document}